\documentclass[conference]{IEEEtran}
\IEEEoverridecommandlockouts
\usepackage{cite}
\usepackage{amsmath,amssymb,amsfonts}
\usepackage{algorithmic}
\usepackage{graphicx}
\usepackage{textcomp}
\usepackage{xcolor}
\usepackage{booktabs} 
\usepackage{siunitx} 
\usepackage{multirow}
\usepackage{subcaption}
\usepackage{caption}
\usepackage{threeparttable}
\usepackage{adjustbox}
\usepackage{siunitx}
\usepackage{booktabs}
\def\BibTeX{{\rm B\kern-.05em{\sc i\kern-.025em b}\kern-.08em
    T\kern-.1667em\lower.7ex\hbox{E}\kern-.125emX}}
\begin{document}

\title{Scaling Laws of Machine Learning for \\ Optimal Power Flow}

\author{
\IEEEauthorblockN{Xinyi Liu}
\IEEEauthorblockA{School of Electric Power Engineering \\
South China University of Technology\\
Guangzhou, China\\
202230160046@mail.scut.edu.cn}
\and
\IEEEauthorblockN{Xuan He}
\IEEEauthorblockA{Information Hub \\
HKUST (Guangzhou)\\
Guangzhou, China\\
xhe085@connect.hkust-gz.edu.cn }
\and
\IEEEauthorblockN{Yize Chen}
\IEEEauthorblockA{Department of Electrical and Computer Engineering\\
University of Alberta \\
Edmonton, Canada\\
yize.chen@ualberta.ca}
}

\maketitle

\begin{abstract}
Optimal power flow (OPF) is one of the fundamental tasks for power system operations. While machine learning (ML) approaches such as deep neural networks (DNNs) have been widely studied to enhance OPF solution speed and performance, their practical deployment faces two critical scaling questions: What is the minimum training data volume required for reliable results? How should ML models' complexity balance accuracy with real-time computational limits? Existing studies evaluate discrete scenarios without quantifying these scaling relationships, leading to trial-and-error-based ML development in real-world applications. This work presents the first systematic scaling study for ML-based OPF across two dimensions: data scale (0.1K-40K training samples) and compute scale (multiple NN architectures with varying FLOPs).  Our results reveal consistent power-law relationships on both DNNs and physics-informed NNs (PINNs) between each resource dimension and three core performance metrics: prediction error (MAE), constraint violations and speed. We find that for ACOPF, the accuracy metric scales as $\text{MAE}_{P_g}(D) = 1.2 D^{-0.39} (R^2 = 0.9)$ with dataset size and $\text{MAE}_{P_g}(C) = 2.1C^{-0.25} (R^2 = 0.6)$ with training compute. These scaling laws enable predictable and principled ML pipeline design for OPF. We further identify the divergence between prediction accuracy and constraint feasibility and characterize the compute-optimal frontier. This work provides quantitative guidance for ML-OPF design and deployments.
\end{abstract}



\section{Introduction}
Optimal power flow (OPF) is a cornerstone optimization problem in power system operations, determining the best generator dispatch for a specified target while satisfying physical and operational constraints\cite{momoh2001review}. Traditional optimization solvers, while guaranteeing solution optimality, face significant computational challenges in modern power grids that demand real-time decision-making under high renewable penetration. To address these challenges, machine learning (ML) approaches, particularly deep neural networks (DNNs), have emerged as promising alternatives, offering magnitude speedups by learning to directly map grid conditions to OPF solutions, and research has demonstrated that well-trained DL models can achieve substantial performance improvements in both solution quality and computational efficiency\cite{Pan2020DeepOPFAF, fioretto2020predicting, Donti2021DC3AL,velloso2021combining}.

Despite these advances, a critical research gap remains: current ML-OPF studies lack a systematic understanding of how model performance scales with key factors---training data and computational budget. This knowledge gap leads to two fundamental questions that practitioners must address through inefficient trial-and-error processes:

\textbf{Data Scale}: How much training data is needed for reliable generalization? Large-scale data collection through AC power flow simulations is resource-intensive and computationally expensive~\cite{10463193}, while it's unknown whether all the OPF performance metrics will benefit from an increase in data volume.





\textbf{Compute Scale}: How should DNN model complexity balance OPF performance requirements with real-time computational constraints? In real-time power system operations, inference requires timely completion \cite{dommel2007optimal}, yet large ML models demand significant computational resources. Determining the optimal trade-off between both compute budget and ML-OPF performance is crucial but remains unquantified in existing literature. And resources may be wasted due to diminishing returns from excessive investment in computing resources.


Current ML-OPF research addresses these questions through isolated experiments on specific test cases with tailored datasets and algorithm designs~\cite{Pan2020DeepOPFAF, fioretto2020predicting, Donti2021DC3AL, velloso2021combining}, lacking a unified framework to systematically analyze how these dimensions interact. Recent work on scaling laws for neural networks, which is pioneered in natural language processing \cite{hestness2017deep, Kaplan2020ScalingLF} and subsequently refined \cite{Hoffmann2022TrainingCL}, has revealed predictable power-law relationships between trained model performance and data/compute resource scales. However, whether such scaling laws exist for physics-constrained optimization problems like OPF, and how they might differ from unconstrained ML tasks in the literature, remains unexplored. 

This work presents the first systematic study of scaling laws for ML-OPF. Through dedicated experiments varying data volumes, system sizes, and computational budgets with performance metrics of prediction error, constraint violations and speed, we discover the answers addressing the above questions with consistent and universal power-law scaling in data size and training compute. We also identify a critical challenge in scaling ML-OPF: prediction accuracy and constraint feasibility improve at different rates as dataset size grows. These results suggest the strategy of data collection and design of training paradigm of ML-OPF approaches, further showing that ensuring ML solution feasibility is a more challenging task than achieving high prediction accuracy for ML-OPF.

\section{Methodology}
To investigate the scaling laws for DL-based OPF, we employ DNN and PINN models as testbeds. By observing the variation in evaluation metrics, including prediction accuracy, physical feasibility, and speed, we aim to discover the scalability concerning training data and computational cost.

\subsection{Deep Learning Testbeds}
 \textbf{DNN Model}:  For both DCOPF and ACOPF problems, the DNN learns a direct mapping. In DCOPF, it learns the active power load vector $\mathbf{P}_d$ to the optimal active power generator dispatch vector $\mathbf{P}_g$ \cite{Pan2019DeepOPFDN}. While in ACOPF, the network maps both active ($\mathbf{P}_d$) and reactive ($\mathbf{Q}_d$) power loads to a vector $\mathbf{y}$ of the primary decision variables: active power generation ($\mathbf{P}_g$) and bus voltage magnitudes ($\mathbf{V}_m$). The network is trained to minimize the Mean Squared Error (MSE) $\mathcal{L}_{\text{DC-DNN}}$ between the predicted dispatch $\hat{\mathbf{P}}_g$ and the ground-truth optimal dispatch $\mathbf{P}_g^*$, and MSE $\mathcal{L}_{\text{AC-DNN}}$ between the predicted solution vector $\hat{\mathbf{y}}$ and the ground-truth optimal solution $\mathbf{y}^*$.

\textbf{PINN Model}: For DCOPF, physics-informed NN (PINN) architecture leverages Karush-Kuhn-Tucker (KKT) optimality conditions as physics constraints~\cite{Nellikkath2021PhysicsInformedNN, chen2022learning}, consisting of two parallel neural networks: one  predicts the generator dispatch $\mathbf{P}_g$, and the other 
predicts the Lagrangian multipliers $\mathbf{\lambda}$ and $\mathbf{\mu}$ associated with the OPF constraints. For ACOPF, the training objective combines the standard supervised MSE loss with a physical penalty term $\mathcal{P}_{\text{phys}}$, of which the gradient is estimated using a zero-order finite difference during backpropagation. Our model directly maps the input to the complete vector of primary decision variables $\hat{\mathbf{y}} = (\hat{\mathbf{P}}_g, \hat{\mathbf{V}}_m)$ to streamline the implementation with DNN and adopts the core mechanism from \cite{Pan2020DeepOPFAF} for the penalty calculation: $\mathcal{P}_{\text{phys}}$ is computed by a custom layer that first extracts the predicted generator setpoints ($\hat{\mathbf{P}}_g$, $\hat{\mathbf{V}}_m^{\text{gen}}$) from the network's output $\hat{\mathbf{y}}$ and feeds them, along with input loads $\mathbf{x} = (\mathbf{P}_d, \mathbf{Q}_d)$, into a full AC Power Flow (ACPF) simulation using \emph{PyPower}. The penalty aggregates all operational constraint violations ($P_g/Q_g$ limits, $V_m$ bounds, and branch flow limits).

\subsection{Evaluation Metrics}

\subsubsection{Prediction Accuracy}
Prediction accuracy is quantified using percentage Mean Absolute Error (MAE) for $N$ samples:
\begin{equation}
\text{MAE}_{\text{var}} = \frac{1}{N} \sum_{i=1}^{N} \frac{|\hat{y}_{i} - y_{i}|}{|\bar{y}| + \epsilon} \times 100\%;
\label{eq:mae}
\end{equation}
where $\hat{y}_i$ and $y_i$ denote predicted and ground-truth values for a specific variable (e.g., $P_g$), $\bar{y}$ is the mean absolute ground truth for that variable, $N$ is the number of test samples, and $\epsilon=10^{-8}$ prevents division by zero. We report the $\text{MAE}$ of $P_g$ for DCOPF and the $\text{MAE}$ of $P_g$ and $V_m$ for ACOPF.

\subsubsection{Physical Feasibility}
For ACOPF, we take the DL predictions $\hat{P}_g$ and $\hat{V}_m$ as setpoints to calculate ACPF in \emph{PyPower}. Then, the violations $\nu_{X}$ for $P_g$, $Q_g$, and $V_m$ are uniformly defined as the absolute amount by which the ACPF-solved value ($X^{\text{pf}}$) exceeds its operational limits ($X^{\min}, X^{\max}$):
\begin{equation}
\nu_{X} = \max(0, X^{\min} - X^{\text{pf}}) + \max(0, X^{\text{pf}} - X^{\max});
\label{eq:viol_limits}
\end{equation}
For DCOPF, the active power violation $\nu_{P_g, j}^{\text{dc}}$ is computed directly from the model prediction ($\hat{P}_{g,j}$) against its limits:
\begin{equation}
\nu_{P_g, j}^{\text{dc}} = \max(0, P_{g,j}^{\min} - \hat{P}_{g,j}) + \max(0, \hat{P}_{g,j} - P_{g,j}^{\max});
\label{eq:dc_pg_vio}
\end{equation}
For both problems, Branch Violation (\%) for each branch $l$ (with limit $S_l^{\max} > 0$) is defined as:
\begin{equation}
\nu_{S_l, \%} = \max\left(0, \frac{S_l^{\text{flow}}}{S_l^{\max}} - 1\right) \times 100\%;
\label{eq:viol_branch_merged}
\end{equation}
where $S_l^{\text{flow}}$ represents the ACPF-solved apparent power for ACOPF, or the magnitude of the DC power flow for DCOPF. For all violation metrics, we report the Mean Max Violation. This is computed by recording the maximum violation per sample, then averaging them over all $N$ test samples.

\subsubsection{Speed}
We evaluate both training time and inference time, while inference time is the sample average time.

\subsection{Scaling Law Discovery}

Based on the above components, we implement and instantiate our scaling law discovery framework:

\begin{itemize}
\item \textbf{Power-Law Characterization.} 
We characterize performance scaling using power-law functions~\cite{hestness2017deep,Kaplan2020ScalingLF}. For a given performance metric $m$ (e.g., MAE) and scaling resource $x$ (e.g., dataset size $D$ or training compute $C$), the relationship is expressed as:
\begin{equation}
m(x) = a \cdot x^{\alpha} (R^2);
\end{equation}
where $a$ is the scaling coefficient and $\alpha$ is the scaling exponent that quantifies scaling efficiency: larger magnitude implies faster improvement. We use non-linear least squares to minimize $\sum (m_i - a \cdot x_i^{\alpha})^2$ over all experimental data points. The coefficient of determination $R^2$ measures the predictability and reliability of the fitted scaling law. Such a fitted law also helps generalize to unseen combination of compute and dataset size.

\item \textbf{Data Scale.} 
We vary the volume of training data to capture the relationship between dataset size $D$ and ML-OPF model performance.
This allows us to predict performance improvements from increased data collection and identify potential data bottlenecks in model training.

\item \textbf{Compute Scale.} 
We quantify training computational cost using floating-point operations (FLOPs)~\cite{amodei2018ai}. For each fully-connected layer with $n_{\text{in}}$ inputs and $n_{\text{out}}$ outputs:
\begin{equation}
\text{FLOPs}_{\text{layer}} = 2 n_{\text{in}} n_{\text{out}} + n_{\text{out}}.
\end{equation}
Total training FLOPs can then be calculated as
\begin{equation}
\text{FLOPs}_{\text{total}} = 3 \times \text{FLOPs}_{\text{forward}} \times N_{\text{train}} \times N_{\text{epochs}};
\end{equation}
where the factor of $3$ accounts for both forward and backward propagation~\cite{hobbhahn2021backward}. We establish empirical scaling laws  adapted from~\cite{hestness2017deep, Kaplan2020ScalingLF} examining architecture scaling (optimal network configurations) and compute budget scaling (diminishing returns with training duration)\cite{Hoffmann2022TrainingCL}.
\end{itemize}

\begin{figure*}
\vspace{-2em}
    \centering
    \includegraphics[width=1.0\linewidth]{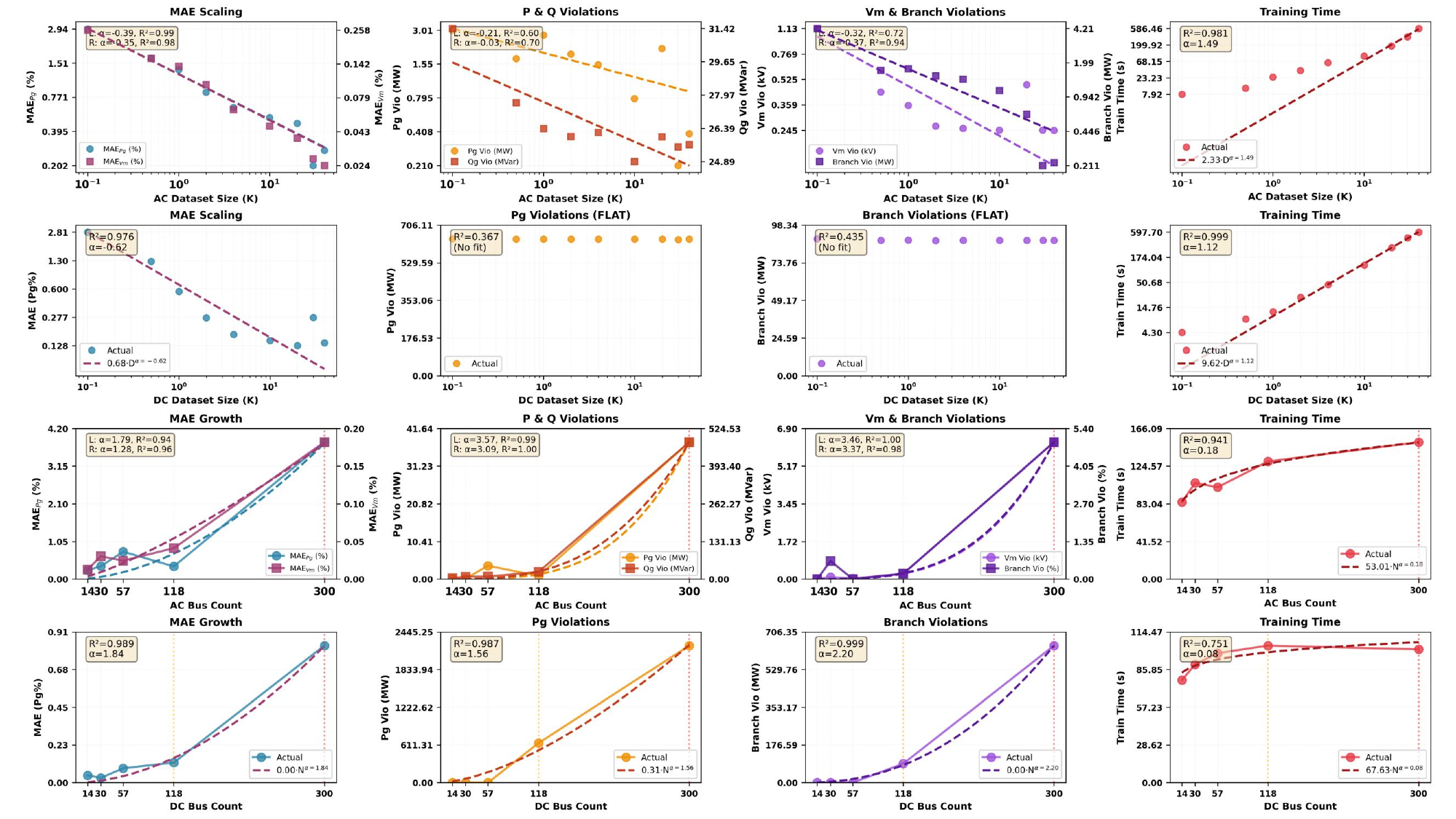}
    \caption{(First two rows) Data and (Last two rows) system scaling. Fitted power law scaling functions are also drawn.}
    \label{fig:data_system_scaling}
\end{figure*}
\vspace{-5pt}
\section{Scaling Law Experiments}

\subsection{Experiment Setups}

We benchmark on IEEE test cases (14-, 30-, 57-,118- and 300-bus), sourced from PGLib-OPF \cite{Babaeinejadsarookolaee2019ThePG}. Both training and held-out testing datasets are generated using \emph{PowerModels.jl} and \emph{Ipopt} along with \emph{Gaussian} perturbation \cite{1425572}. We input 50,000 load samples per case, retaining only the converged and feasible cases as resulting data.



For \texttt{Data Scaling} case, we use 118-bus system and NN architecture of [128,128] for DCOPF and of [200,200] for ACOPF, with training samples from 0.1K to 40K. For \texttt{Compute Scaling} case,  we focus on the 118-bus system, 10K samples and evaluate how architecture scaling of 28 configurations (1--4 layers, 32--1024 neurons, ReLU activation function, 1K epochs); and compute budget scaling of varying training durations (0.06K--5K epochs) under different NN architectures would affect testing performance.

All simulations are conducted using AMD Ryzen 7 5800H CPU with no GPU acceleration.
\vspace{-2pt}
\subsection{Results for \texttt{Data Scaling}}

\begin{table}[bth]
\vspace{-2pt}
\centering
\caption{\texttt{Data Scaling} results (1K epochs, 118-bus).}
\label{tab:dnn_data_scaling}
\small
\begin{tabular}{lrrrrr}
\toprule
\textbf{Metric} & \textbf{0.1K} & \textbf{1K} & \textbf{10K} & \textbf{40K} & \textbf{Impro.} \\
\midrule
\multicolumn{6}{c}{DCOPF} \\
\midrule
MAE (Pg\%) & 2.82 & 0.56 & 0.15 & 0.14 & 95\% ↓ \\
Pg Viol (MW) & 641.9 & 640.5 & 640.3 & 641.0 & 0\% (flat) \\
Branch Viol (\%) & 89.4 & 88.5 & 88.4 & 88.5 & 1\% (flat) \\
\midrule
\multicolumn{6}{c}{ACOPF} \\
\midrule
MAE (Pg\%) & 2.94 & 1.33 & 0.52 & 0.27 & 91\% ↓ \\
MAE (Vm\%) & 0.26 & 0.14 & 0.05 & 0.02 & 92\% ↓ \\
Pg Viol (MW) & 3.01 & 2.73 & 0.78 & 0.39 & 87\% ↓ \\
Qg Viol (MVar) & 31.4 & 26.4 & 24.9 & 25.7 & 18\% ↓ \\
Branch Viol (\%) & 4.21 & 1.75 & 1.08 & 0.26 & 95\% ↓ \\
\bottomrule
\end{tabular}
\end{table}

\subsubsection{Accuracy Follows Steep Power-Law Scaling}

All testing errors exhibit diminishing patterns as training dataset grows. Both DCOPF and ACOPF exhibit strong power-law relationships between prediction accuracy and training data size:\vspace{-15pt}

\begin{align}
\text{DCOPF: } \text{MAE}(D) &= 0.7 \cdot D^{-0.624} \quad (R^2 = 0.976) \\
\text{ACOPF: } \text{MAE}_{P_g}(D) &= 1.2 \cdot D^{-0.389} \quad (R^2 = 0.992) \\
\text{MAE}_{V_m}(D) &= 0.12 \cdot D^{-0.349} \quad (R^2 = 0.985)\vspace{-20pt}
\end{align}

As shown in Table \ref{tab:dnn_data_scaling}, with 400× more data, DCOPF MAE improves 95\% and ACOPF MAE improves 91\% (Pg), showing that prediction accuracy scales predictably in both cases. 


\subsubsection{Violations Exhibit Structural Scaling Bottleneck}

Compared to the accuracy metric, the \texttt{Data Scaling} in constraint violations shows weakness and inconsistency. For DCOPF case, both $P_g$ violations and branch violations remain flat across all data scales. This is probably inherent from the fact that DCOPF's linear approximation creates errors that no additional amount of training data can overcome. While for the ACOPF case, we fit the following power law: \vspace{-15pt}

\begin{align}
\text{Pg Vio}(D) &= 1.93 \cdot D^{-0.207} \quad (R^2 = 0.600) \\
\text{Qg Vio}(D) &= 27.6 \cdot D^{-0.030} \quad (R^2 = 0.700) \\
\text{Branch Vio}(D) &= 1.74 \cdot D^{-0.369} \quad (R^2 = 0.938)
\end{align}
Among them, branch violations are well-captured by the loss function and scale almost as fast as accuracy metrics, while generator constraints especially reactive power limits remain the prediction bottleneck. The overall trend of $P_g$ violations show 87\% improvement (3.01 → 0.39 MW), suggesting that $P_g$ violations benefit from \texttt{data scaling} but are more sensitive to optimization dynamics than accuracy metrics.




\begin{table}[h]
\centering
{\fontsize{6.5}{15}\selectfont
\setlength{\tabcolsep}{4.5pt}
\caption{DNN vs PINN: Accuracy-feasibility trade-off (60 epochs, 118-bus).}
\label{tab:dnn_pinn_tradeoff}
\small
\begin{tabular}{llrrrr}
\toprule
\textbf{Problem} & \textbf{Method} & \textbf{Data} & \textbf{MAE} & \textbf{Viol} & \textbf{Ratio} \\
\midrule
\multirow{4}{*}{DCOPF} & DNN & 1K & 2.21(Pg) & 633.4(Pg) & -- \\
& PINN & 1K & 4.56 (Pg) & 0.94 (Pg) & 2.1${\uparrow}$ / 674${\downarrow}$ \\
\cmidrule{2-6}
& DNN & 40K & 0.39 (Pg) & 640.4 (Pg) & -- \\
& PINN & 40K & 4.73 (Pg) & 0.88 (Pg) & 12${\uparrow}$ / 728${\downarrow}$ \\
\midrule
\multirow{8}{*}{ACOPF} & DNN  & 1K & 3.22 (Pg) & 2.29(Pg) & -- \\ 
& & & 0.25(Vm) & 38.1(Qg) & \\
& PINN & 1K & 4.03  (Pg) & 1.34 (Pg) & 1.3${\uparrow}$ / 1.7${\downarrow}$ \\
& & & 0.35 (Vm)& 6.0 (Qg)& 1.4${\uparrow}$ / 6.4${\downarrow}$ \\
\cmidrule{2-6}
& DNN & 40K & 0.57 (Pg) & 0.84 (Pg) & -- \\
& & & 0.06  (Vm)& 25.7 (Qg) & \\
& PINN & 40K & 2.29 (Pg) & 0.006 (Pg) & 4.0${\uparrow}$ / 140${\downarrow}$ \\
& & & 0.22 (Vm) & 9.1 (Qg) & 3.7${\uparrow}$ / 2.8${\downarrow}$ \\
\bottomrule
\multicolumn{6}{l}{\footnotesize MAE: Pg (\%) for DCOPF; Pg/Vm (\%) for ACOPF.} \\
\multicolumn{6}{l}{\footnotesize Viol: Pg (MW) for DCOPF; Pg (MW)/Qg (MVar) for ACOPF.}\\
\multicolumn{6}{l}{\footnotesize Ratio: left, MAE; right, Viol.}
\end{tabular}}
\end{table}

\subsubsection{Feasibility-Accuracy Trade-off in DL Structure} 
In the DCOPF case, Table \ref{tab:dnn_pinn_tradeoff} shows that a \textbf{728× improvement} is achieved by PINN on the violation metric compared to DNN. This indicates that violation scaling by adding data is inherently limited by DNN's architecture, while PINN can yield 2-3 orders of magnitude improvement. In contrast, for the accuracy metric, PINN's MAE is 12× worse than DNN (4.73\% vs 0.39\%),  suggesting that enforcing non-linear AC physics (e.g., KKT conditions) may conflict with PINN training objectives.

In the ACOPF case, PINN achieves steeper violation scaling. PINN dramatically improves $P_g$ violations but struggles with $Q_g$ violations. Critically, PINN with 4K samples ($P_g$ viol = 0.02 MW) outperforms DNN with 40K samples (0.84 MW), achieving 42× better $P_g$ violation control with 10× less data. This quantifies that specificed NN model could overcome data scaling limitations for active power constraints.



This feasibility-accuracy trade-off reflects that DNN's MSE loss prioritizes value prediction; PINN's physics penalty prioritizes active power constraint satisfaction at the expense of prediction accuracy. Neither eliminates the trade-off---they occupy opposite ends of the Pareto frontier\cite{Rohrhofer2021DataVP}.





\begin{table}[h]
\centering
{
\setlength{\tabcolsep}{1.3pt}
\caption{Performance comparison on varying IEEE test cases}
\label{tab:system_scaling}
\begin{tabular}{lcccccc}
\toprule
\textbf{Problem} & \textbf{Method} & \textbf{14} bus & \textbf{118} bus & \textbf{300} bus & \textbf{Growth} & \textbf{Ratio} \\
& &  &  && \textbf{(118→300)} & \textbf{(300/118)} \\
\midrule
\multicolumn{7}{c}{\textit{DNN Performance (1K epochs, 10K samples)}} \\
\midrule
\multirow{2}{*}{DCOPF} & MAE (\%) & 0.04 & 0.12 & 0.82 & +0.70\% & 6.8× \\
& Pg Viol (MW) & 0.02 & 641.2 & 2223 & +1582  & 3.5× \\
\midrule
\multirow{2}{*}{ACOPF} & MAE Pg (\%) & 0.24 & 0.36 & 3.82 & +3.46\% & 10.6× \\
& Qg Viol (MVar) & 3.2 & 25.2 & 477 & +452  & 18.9× \\
\midrule
\multicolumn{7}{c}{\textit{DNN vs PINN Comparison (60 epochs, 1K samples)}} \\
\midrule
{DCOPF} & DNN MAE (\%) & 0.28 & 2.21 & 4.07 & +1.86\% & 1.8× \\
\textit{MAE} & PINN MAE (\%) & 3.35 & 4.37 & 7.22 & +2.85\% & 1.7× \\
\midrule
{DCOPF} & DNN Viol (MW) & 0.18 & 643.2 & 2224 & +1581  & 3.5× \\
\textit{Pg Viol} & PINN Viol (MW) & 0.00 & 1.29 & 10.6 & +9.3  & 8.2× \\
\midrule
{ACOPF} & DNN MAE (\%) & 1.34 & 2.70 & 8.01 & +5.31\% & 3.0× \\
\textit{MAE Pg} & PINN MAE (\%) & 2.76 & 4.03 & 25.0 & +21.0\% & 6.2× \\
\midrule
{ACOPF} & DNN Viol (MVar) & 3.8 & 30.0 & 477 & +447  & 15.9× \\
\textit{Qg Viol} & PINN Viol (MVar) & 1.7 & 6.0 & 199 & +193  & 33.2× \\
\bottomrule
\end{tabular}}
\end{table}

\subsection{Results for \texttt{Compute Scaling}}

\subsubsection{Varying \texttt{Compute Scaling} Laws}

\begin{figure}
    \centering
    \includegraphics[width=1.0\linewidth]{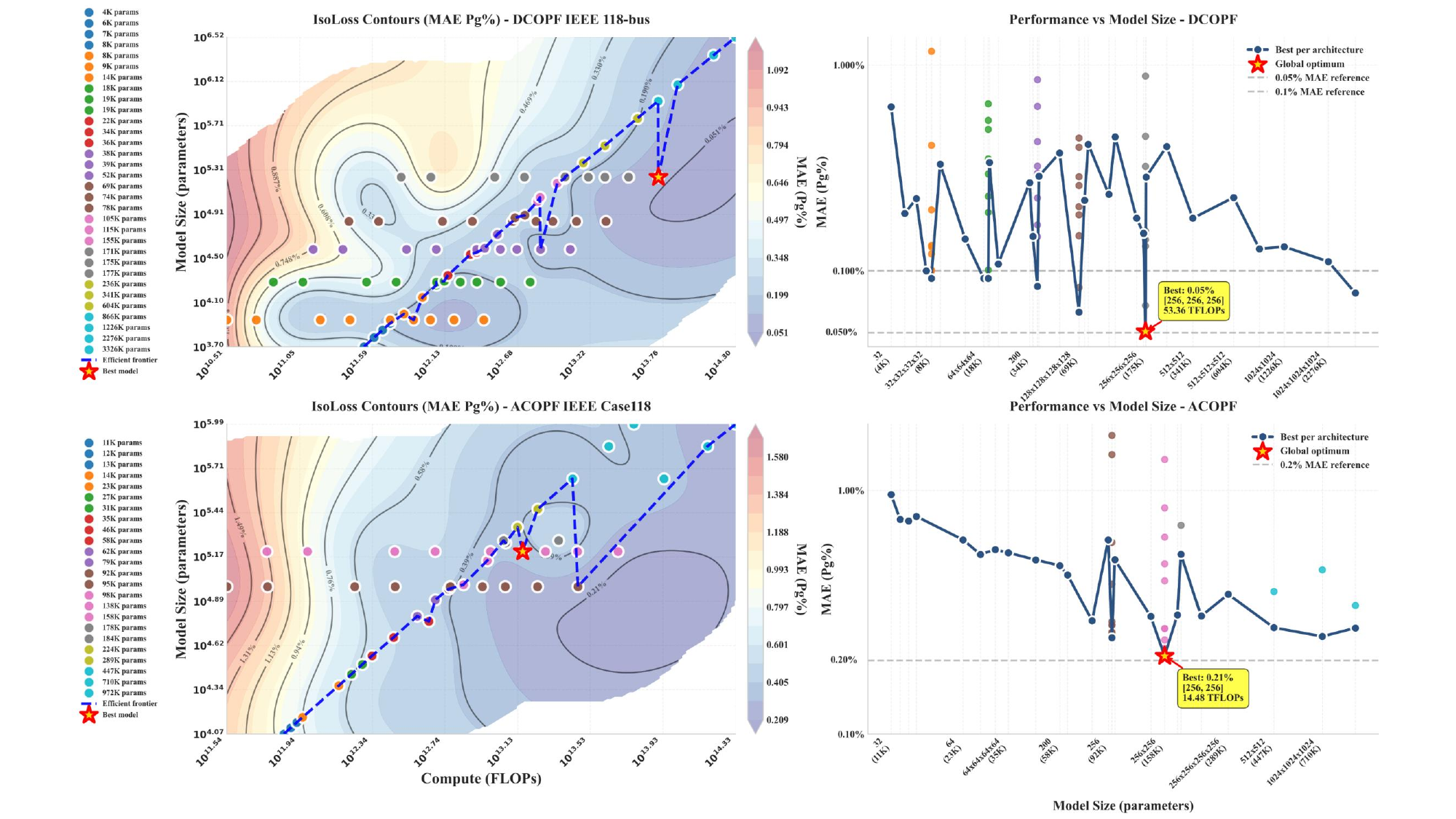}
    \caption{\texttt{Compute Scaling} on 118-bus system. IsoLoss contours show prediction error (color) versus training compute (FLOPs) and model size (Parameters). Dashed line: efficiency frontier. (Right) Best performance per architecture (blue) and global optimum (red star); error measured as MAE for $P_g$.}
    \label{fig:compute_scale}
\end{figure}

For DCOPF on IEEE 118-bus system, we evaluate model configurations at a fixed batch size of 128 with compute budgets from 0.032 to 201.8 TFLOPs. \texttt{Compute Scaling} law for DCOPF is:
\begin{equation}
\text{MAE}_{P_g}(C) = 0.296 \cdot C^{-0.191}, \quad R^2 = 0.259
\label{eq:dcopf_compute}
\end{equation}

This indicates a 10-fold increase in training compute reduces prediction error by 35.6\%. However, the modest $R^2$ value of 0.259 reveals substantial variance, suggesting that architecture choice and optimization dynamics introduce additional complexity beyond simple \texttt{Compute Scaling}. Also, it is observed that for DCOPF case, the 3-layer NN architecture emerges as the optimal, outperforming both shallower (2-layer) and deeper (4-layer) models with same TFLOPS.


For ACOPF problem, we evaluate configurations at batch size 200, yielding compute budgets from 0.35 to 211.7 TFLOPs. The empirical scaling laws for ACOPF are:
\begin{align}
\text{MAE}_{P_g}(C) &= 2.065 \cdot C^{-0.250}, \quad R^2 = 0.586 \label{eq:acopf_pg_compute} \\
\text{MAE}_{V_m}(C) &= 0.188 \cdot C^{-0.216}, \quad R^2 = 0.632 \label{eq:acopf_vm_compute}
\end{align}

The ACOPF scaling exponent for active power ($\alpha_C = -0.250$) is 31\% stronger than DCOPF ($\alpha_C = -0.191$). However, ACOPF's optimal depth is 2 layers (for $P_g$) compared to DCOPF's 3 layers, given fixed TFLOPS. We attribute this to the structured nature of AC nonlinearities: power flow equations contain quadratic voltage terms and trigonometric phase relationships that wide shallow networks can directly approximate. In contrast, DCOPF's linear system benefits from an additional layer to capture higher-order interactions in the linearized power balance equations.


\begin{table}[t]
\caption{Diminishing returns of ML ([256,256]) for ACOPF case, \texttt{Compute Scaling.}}
\label{tab:acopf_compute}
\centering
\begin{tabular}{l c c c c c} 
\toprule
Epochs & MAE (Pg\%) &TFLOPs & $\Delta$TFLOPs &$\Delta$MAE & \shortstack[c]{Efficiency \\ ($\frac{\Delta\text{MAE}}{\Delta\text{TFLOPs}}$)} \\
\midrule

60 & 1.34 &0.58& - & - & - \\
300 & 0.42 &2.90& 2.32 &-0.92& -4e-1 \\
1000 & 0.27 &9.65& 6.75 & -0.15 & -2.2e-2 \\
1500 & 0.21 &14.48& 4.83 & -0.06 & -1.2e-2 \\
5000 & 0.22 &48.26& 33.78 & 0.01 & +3.0e-4 \\
\bottomrule
\end{tabular}
\end{table}


Table~\ref{tab:acopf_compute} reveals diminishing returns in extended training compute, aligning with the compute-optimal training paradigm in \cite{Hoffmann2022TrainingCL}. Results show that it is appropriate to implement early stopping of training and reallocating saved computational budget to train larger models with fewer epochs. For instance, extending training to 5,000 epochs while increasing compute to 48.26 TFLOPs fails to improve performance. Further, compared to the initial training phase, where marginal efficiency reaches 0.219 \%/TFLOP, the 1,500–5,000 epoch increase sees marginal efficiency plummet to 0.000477 \%/TFLOP.  

\subsubsection{Implications for ML-OPF} Our observation on compute-optimal shallow-wide  networks for ACOPF indicates that  given fixed compute budget,  neural architecture search of OPF should not blindly adopt deeper structures.  The scaling exponents we observe ($\alpha_C = -0.191$ for DCOPF, $-0.250$ for ACOPF) are 3.8$\times$ to 5$\times$ stronger than language models ($\alpha_C \approx -0.050$ in \cite{Kaplan2020ScalingLF}), indicating that ML-based OPF solvers exhibit steeper learning curves with respect to compute. It also suggests that OPF problems under sufficiently engineering complex enable NN models to extract significant performance improvements from additional training.

\section{Discussions and Conclusion}
This paper presents the first systematic study of the scaling laws for Machine Learning-based Optimal Power Flow (ML-OPF),  examining performance as a function of training data volume and computation budget. The research clearly reveals that ML-OPF performance, particularly prediction accuracy (MAE), follows a highly predictable power-law relationship with both data size and computation, providing a quantitative basis for data collection and training paradigms in ML-OPF algorithm design. However, we discover the significant divergence observed between prediction accuracy and physical constraint feasibility. As resources increase, prediction accuracy (MAE) improves steadily and rapidly, but the violation of physical constraints (especially reactive power) improves much more slowly, becoming the critical bottleneck for practical deployments. Specially for DCOPF problem, increasing data scale (from 100 to 40,000 samples) or compute scale (from 60 to 5,000 epochs) have much less impacts on improving its solution feasibility. 

In the future work, we would like to extend the scaling law analysis for OPF to broader ML algorithms, and we will also investigate the effects of underlying power network size, regularization on ML-OPF performance.

\bibliographystyle{IEEEtran}
\bibliography{reference}

\end{document}